\documentclass[times, review, 10pt]{elsarticle}
\usepackage[caption=false,font=normalsize,labelfont=sf,textfont=sf]{subfig}
\usepackage{booktabs} 
\usepackage{capt-of}
\usepackage{color}
\definecolor{cvprblue}{rgb}{0.21,0.49,0.74}
\usepackage[pagebackref,breaklinks,colorlinks,citecolor=cvprblue,hypertexnames=false]{hyperref}
\usepackage{algorithm}
\usepackage{listings}
\definecolor{moco}{RGB}{33, 130, 129}
\definecolor{myRed}{RGB}{195,10,10}
\definecolor{myGreen}{RGB}{55,149,73}
\usepackage{multirow}
\usepackage{xspace}
\usepackage{etoolbox}
\makeatletter
\AfterEndEnvironment{algorithm}{\let\@algcomment\relax}
\AtEndEnvironment{algorithm}{\kern2pt\hrule\relax\vskip3pt\@algcomment}
\let\@algcomment\relax
\newcommand\algcomment[1]{\def\@algcomment{\footnotesize#1}}
\renewcommand\fs@ruled{\def\@fs@cfont{\bfseries}\let\@fs@capt\floatc@ruled
  \def\@fs@pre{\hrule height.8pt depth0pt \kern2pt}%
  \def\@fs@post{}%
  \def\@fs@mid{\kern2pt\hrule\kern2pt}%
  \let\@fs@iftopcapt\iftrue}
\makeatother

\usepackage{graphicx}
\usepackage[dvipsnames]{xcolor}
\definecolor{citecolor}{HTML}{0071bc}
\definecolor{Highlight}{HTML}{39b54a}  % green

\usepackage{xcolor}

\usepackage{amsmath,amsfonts}
\usepackage{algorithmic}
\usepackage{algorithm}
\usepackage{array}
\usepackage[capitalize]{cleveref}

\newcommand{\revise}[1]{\textcolor{black}{#1}}

\usepackage{subfig}
\usepackage{amssymb}
\usepackage{amsmath}
\usepackage{amsthm}

\usepackage{lineno}
\usepackage{cleveref}

\usepackage{xspace}
\makeatletter
\DeclareRobustCommand\onedot{\futurelet\@let@token\@onedot}
\def\@onedot{\ifx\@let@token.\else.\null\fi\xspace}

\makeatother
\usepackage{booktabs} 
\journal{Pattern Recognition}

\begin{document}

\begin{frontmatter}



    \title{Separators for Enhancing Autoregressive Pretraining in Vision Mamba}


    %% Author name
    \author[label1]{Hanpeng Liu} \ead{hanpengliu@hust.edu.cn}
    \author[label1]{Zidan Wang}
    \ead{m202574073@hust.edu.cn}
    \author[label1]{Shuoxi Zhang}
    \ead{zhangshuoxi@hust.edu.cn}
    \author[label1]{Kaiyuan Gao}
    \ead{im_kai@hust.edu.cn}
    \author[label1]{Kun He\corref{cor1}\fnref{label2}}
    \ead{brooklet60@hust.edu.cn}
    \cortext[cor1]{Corresponding author}
    \fntext[label2]{This work is supported by National Natural Science Foundation of China (U22B2017) and International Cooperation Foundation of Hubei Province, China (2024EHA032).}

    \affiliation[label1]{organization=School of Computer Science and Technology, addressline={Huazhong University of Science and Technology},
        city={Wuhan},
        postcode={430074},
        state={Hubei},
        country={China}}
    % \affiliation{organization={},%Department and Organization

    %% Abstract
    \begin{abstract}
        The state space model Mamba has recently emerged as a promising paradigm in computer vision, attracting considerable attention for its efficient handling of long-sequence tasks. Its inherent causal structure makes it particularly well suited for autoregressive pretraining. However, existing autoregressive pretraining methods in vision are largely limited to short-sequence settings and may not fully use Mamba's capacity to model longer contexts. To investigate this setting, we introduce SeparaTors for AutoRegressive pretraining (STAR), a new autoregressive pretraining method for Vision Mamba that explicitly marks the boundaries between different images. STAR increases the patch-token sequence length from 144 to 640 by packing four images and four separator clusters. This is approximately $4.4\times$ the ARM patch-token sequence length. The increase is achieved without changing the resolution of any individual image: we use $192\times192$ inputs for autoregressive pretraining and $224\times224$ inputs for downstream classification fine-tuning. With this long-sequence pretraining scheme, STAR-B achieves 83.5\% EMA top-1 accuracy on ImageNet-1K after 1,600 epochs of pretraining. The learned representation also transfers beyond in-distribution classification: compared with ARM, STAR-B improves COCO box AP from 46.11 to 46.84 and mask AP from 40.74 to 41.45, while raising the mean top-1 accuracy across five ImageNet robustness benchmarks from 55.1\% to 56.8\%. Under the evaluated four-image setting, these results indicate that separator-based long-sequence pretraining improves recognition robustness and dense visual prediction relative to ARM.
    \end{abstract}

    %% Graphical abstract
    % % \begin{figure*} [t!]
    % % 	\centering
    % % 	\includegraphics[width=1.0\textwidth]{PVP.pdf}
    % % 	\caption{\textbf{The proposed P2P architecture.} Each batch of samples $B$ are augmented and  divided into two parts, denoted as $B'$ and $B''$. In the $t$-th iteration ($t \ge 1$), not only new samples $B_{t}''$ are sampled in computing loss $\mathcal{L}_{EMP}$,  but also  data $B_{t-1}'$ enhanced in the $t-1$ step are encoded by $f(\theta_{t})$ and $f(\xi_{t})$ to obtain embedding $Z_{t-1}'$. Additionally, the loss $\mathcal{L}_{P2P}$ is calculated by comparing the past encoding $Z_{t-1}''$ with the current encoding $Z_{t-1}'$.
    % %     }
    % %     \label{PVP} 
    % % \end{figure*}
    % \includegraphics[width=1.0\textwidth]{abstract.pdf}

    \begin{highlights}
        % \item We are the first to balance both training efficiency and performance, significantly boosting model performance in the one-epoch training regime.
        % \item We propose an innovative P2P (Past-to-Present) smoothing framework. This framework leverages the model’s previous outputs as supervisory signals and distributes multiple patches of the same image across different batches for training, significantly improving patch utilization efficiency and addressing the efficiency bottleneck caused by the excessive use of cropped patches in previous methods.
        % \item We conduct extensive experiments and achieve excellent performance on multiple standard datasets, and our method also shows better generalization compared to other self-supervised learning baselines.
        \item STAR enables long-sequence autoregressive pretraining for Vision Mamba.
        \item Cluster-level separators enable multiple images to be modeled in one sequence.
        \item STAR-B reaches 83.5\% EMA top-1 accuracy after 1,600 pretraining epochs.
        \item STAR improves ARM on COCO and five ImageNet robustness benchmarks.
    \end{highlights}

    %% Keywords

\end{frontmatter}



\section{Introduction}
\label{sec:intro}
State space models~\cite{kalman1960new} have recently attracted substantial interest in deep learning, particularly in natural language processing. Mamba~\cite{Mamba} introduces input-dependent selective state spaces while retaining linear complexity with respect to sequence length, making it efficient for long-context modeling. This property has motivated the development of Vision Mamba models for image classification~\cite{Vim,mambar}, object detection~\cite{VMamba,Mambaout}, and semantic segmentation~\cite{EfficientVMamba,Mamba-UNet,SegMamba}.

\begin{figure}[t]
    \centering
    \includegraphics[width=0.95\linewidth]{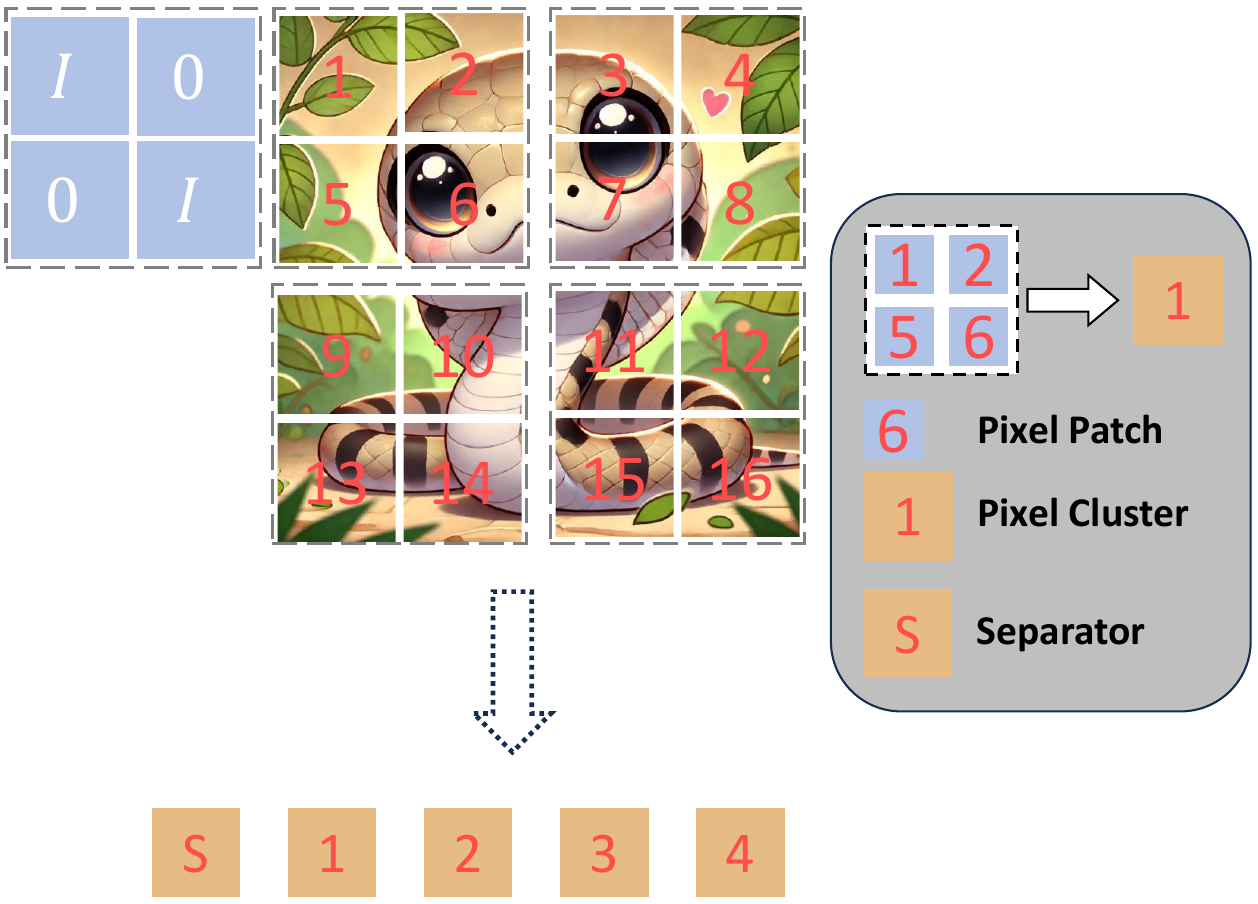}

    \caption{Each image is divided into non-overlapping patches. Multiple spatially adjacent patches are grouped into a cluster. Our STAR adds an input-space separator cluster at the beginning of each image. For the Identity separator, the diagonal positions are all-ones vectors, and all remaining positions are all-zeros vectors.}
    \label{fig:token}
\end{figure}

Self-supervised pretraining for Vision Mamba remains comparatively underexplored. Contrastive learning~\cite{He0WXG20,simclr} and masked image modeling~\cite{BEiT,MAE,simmim} have achieved strong results with convolutional neural networks~\cite{ConvNet} and Vision Transformers~\cite{ViT,crossvit}. However, directly applying these paradigms to Mamba may not fully exploit its causal and recurrent structure. ARM~\cite{ARM} and MAP~\cite{MAP} indicate that autoregressive pretraining is particularly compatible with Vision Mamba~\cite{Vim,VMamba}.

\begin{figure*}[t]
    \centering
    \includegraphics[width=0.97\linewidth]{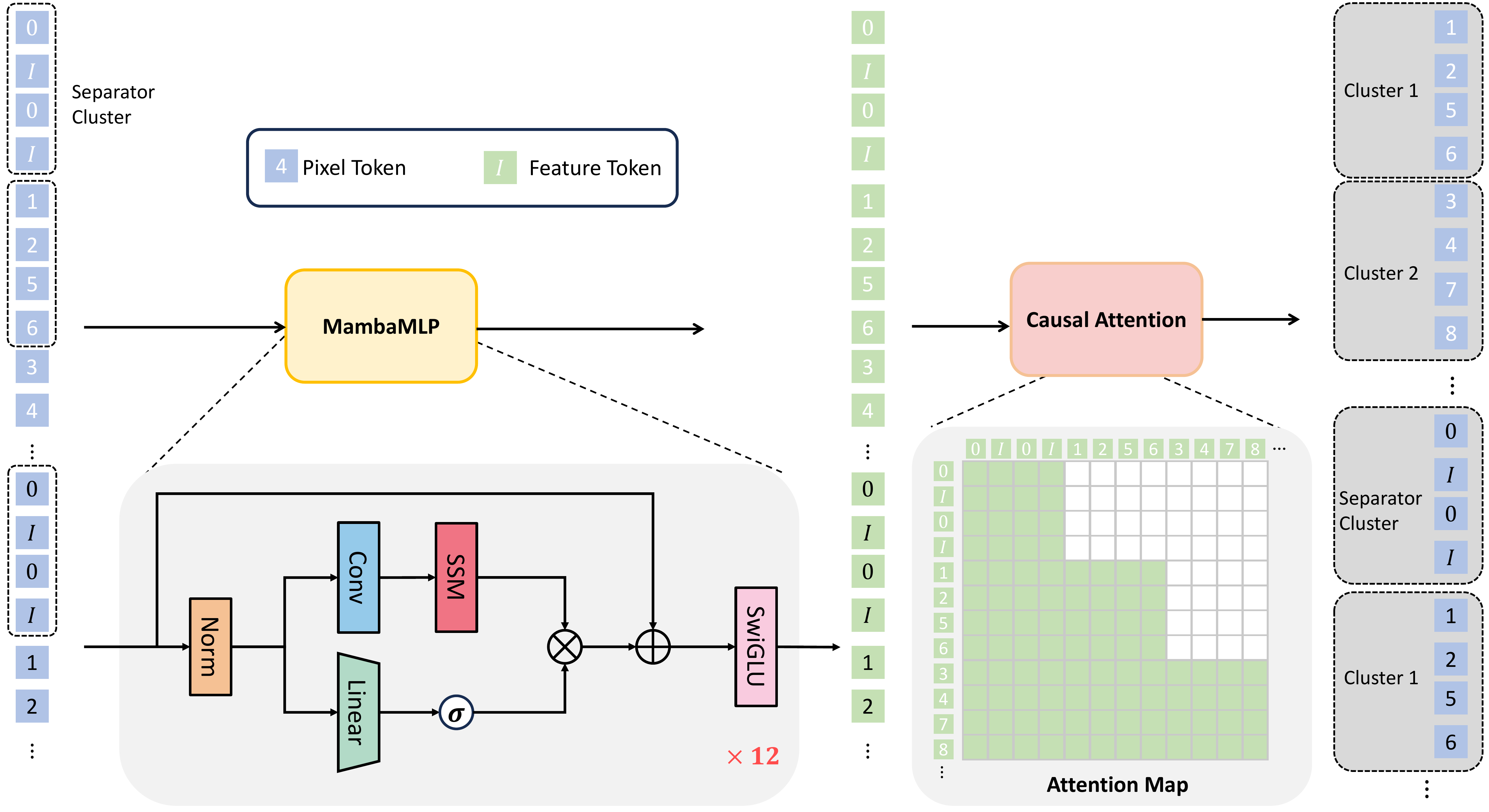}

    \caption{\textbf{Our STAR architecture.} During pretraining, we divide images into patches and group spatially adjacent patches into clusters. We then reorder the patches using a cluster-priority scanning method. Our STAR places a separator cluster at the start of the content clusters of each image and concatenates the resulting per-image sequences into a single long patch-token sequence. This long sequence is fed into a MambaMLP with causal properties. The resulting feature tokens are input to a decoder with causal attention. The attention map of the decoder is bidirectional within clusters and unidirectional between clusters. This allows our STAR to perform autoregression at the granularity of cluster prediction units. The last content cluster of each non-final image predicts the shared separator of the subsequent image, while the last content cluster of the final image predicts an additional terminal separator target.}
    \label{fig:star}
\end{figure*}

Autoregressive pretraining~\cite{XLNet,llama2,iGPT} learns to predict each element from its preceding context. Natural language is inherently sequential, which supports autoregressive models such as GPT-4~\cite{gpt4} and Llama 2~\cite{llama2}. Images, by contrast, are two-dimensional and require an imposed prediction order. Visual autoregressive methods such as iGPT~\cite{iGPT}, AIM~\cite{AIM}, and ARM~\cite{ARM} predict pixels, patches, and clusters of patches, respectively. These methods nevertheless construct relatively short sequences from individual images. This design creates a mismatch with the long-sequence modeling capability of Mamba~\cite{Mambaout}.

In this paper, we introduce \textit{\textbf{S}epara\textbf{T}ors for \textbf{A}uto\textbf{R}egressive pretraining} (\textbf{STAR}), a cluster-level multi-image separator formulation for autoregressive representation pretraining with Vision Mamba. The basic concept is illustrated in~\cref{fig:token}. STAR places a shared, image-agnostic separator cluster before each image, allowing multiple unrelated images to be modeled as a single long-sequence task while preserving explicit image boundaries. Specifically, each image is divided into non-overlapping patches, which are then grouped into clusters of spatially adjacent patches. The separator is created directly in the encoder input space, occupies the same number of positions as a content cluster, and therefore bypasses the patch embedding layer. For the Identity separator used in the final configuration, the diagonal positions are all-ones vectors, and all remaining positions are all-zeros vectors. Then, we model multiple images with separators as a single long sequence following a row-major order, as depicted in~\cref{fig:star}. In this long sequence, each image contributes one separator cluster followed by its content clusters. Finally, this long sequence is fed into a causal encoder alongside a lightweight decoder, which sequentially predicts each cluster. Notably, the encoder is a unidirectional scanning Vision Mamba, while the decoder is a transformer with causal attention~\cite{CausalAttention}. As shown in~\ref{fig:star}, the attention map is bidirectional within clusters and unidirectional between clusters. This design extends the single-image ARM formulation to multi-image sequences and studies whether longer packed contexts benefit Vision Mamba. In our work, we explore variations in the types, values, positions, and numbers of separators. After 1,600 epochs of autoregressive pretraining, our STAR-B model achieves an accuracy of $83.5\%$ on ImageNet-1K.

Our main contributions are summarized as follows:
\begin{itemize}
    \setlength{\itemsep}{0pt}
    \setlength{\parsep}{0pt}
    \setlength{\parskip}{0pt}
    \item  We introduce a cluster-level separator for multi-image autoregressive pretraining with Vision Mamba, enabling unrelated images to form one long sequence with explicit boundaries.

    \item  We formulate cluster-level cross-image autoregression with a shared separator, including a terminal separator target that provides an outgoing next-cluster objective for the final content cluster.

    \item   STAR-B achieves 82.9\% EMA top-1 accuracy on ImageNet-1K after 300 epochs of pretraining and consistently outperforms ARM on COCO object detection and instance segmentation as well as five ImageNet robustness benchmarks, demonstrating transfer beyond in-distribution classification.

\end{itemize}

\section{Related Work}
\label{sec:related work}

\paragraph{Vision Mamba Architectures}
The success of state space models in long-range sequence modeling has motivated their extension from language to vision~\cite{Mamba,Mehta0CN23}. Vim~\cite{Vim} builds a generic visual backbone with bidirectional Mamba scans and linear sequence complexity, while VMamba~\cite{VMamba} captures two-dimensional structure through cross-scanning. Subsequent variants explore sparse scanning~\cite{EfficientVMamba}, architectural simplification~\cite{Mambaout}, and register tokens~\cite{mambar}.

Whereas early Vision Mamba models are largely supervised, ARM~\cite{ARM} shows that autoregressive pretraining suits Mamba's unidirectional recurrence and introduces MambaMLP for token and channel mixing. MAP~\cite{MAP} combines masking and autoregressive prediction for hybrid Mamba--Transformer models. STAR follows ARM and adopts MambaMLP, but reorganizes the input sequence to exploit Mamba's long-context capability rather than modifying the architecture or objective.

\paragraph{Autoregressive Visual Representation Learning}
Autoregressive visual pretraining predicts image content under a chosen factorization order. iGPT~\cite{iGPT} models sequences of discretized pixels, while later methods operate on patches or semantic features. SAIM~\cite{SAIM} uses stochastic patch permutations, D-iGPT~\cite{DiGPT} adds semantic-token prediction, and AIM~\cite{AIM} predicts continuous patch values at scale. DARL~\cite{DARL} combines patch prediction with denoising, whereas AIMv2~\cite{AIMv2} jointly predicts image patches and text tokens. These methods mainly improve targets, objectives, or scalability; STAR instead constructs longer visual contexts from existing images without increasing spatial resolution.

\paragraph{Visual Serialization and Group-wise Prediction}
Multi-image packing predates STAR. NaViT~\cite{NaViT} packs patches for efficient, resolution-flexible Transformer training but masks interactions between images. LVM~\cite{LVM} concatenates discrete image tokens into visual sentences bracketed by boundary tokens for generative pretraining. In contrast, STAR uses a cluster-level separator as both an input boundary and an autoregressive target, propagates recurrent states across images, and learns transferable representations from continuous patch clusters.

Because images lack a unique causal order, prior work has also redesigned serialization and prediction granularity within individual images. RandSAC~\cite{HuaTRRZS23} autoregressively models randomly ordered hierarchical segments, and SemAIM~\cite{SemAIM} prioritizes semantically informative patches. XTRA~\cite{XTRA} uses block-causal attention to model spatial blocks autoregressively while allowing bidirectional interaction within each block. STAR instead combines clusters with explicit separators to form a continuous multi-image sequence.

Long-context autoregressive modeling has been studied in naturally continuous data: Toto~\cite{Toto} models sequences of video frames and images, and TokenUnify~\cite{TokenUnify} applies Mamba-based predictive coding to volumetric microscopy. STAR instead connects unrelated natural images with an image-agnostic separator, preserving identifiable boundaries while extending cluster-level prediction to multi-image long sequences.

\paragraph{Other Self-supervised Learning Paradigms}
Other major self-supervised paradigms include contrastive learning, represented by SimCLR~\cite{simclr}, MoCo~\cite{He0WXG20}, MoCo v3~\cite{mocov3}, and DINO~\cite{CaronTMJMBJ21}, and masked image modeling, including BEiT~\cite{BEiT}, MAE~\cite{MAE}, SimMIM~\cite{simmim}, and MaskFeat~\cite{00050XWYF22}. However, ARM~\cite{ARM} and MAP~\cite{MAP} indicate that conventional masked modeling may not fully exploit causal Mamba's recurrent, unidirectional structure. We therefore focus on autoregressive pretraining with long multi-image sequences.
\section{Method}

\subsection{Mamba Preliminaries}
Mamba~\cite{Mamba,Vim} builds on structured state space models (SSMs) while retaining linear computational complexity in the sequence length. We first introduce the classical SSM formulation and then distinguish it from the selective SSM used by Mamba. A classical structured SSM can be formulated as a linear time-invariant (LTI) system that maps an input $x(t) \in \mathbb{R}$ to an output $y(t) \in \mathbb{R}$ through a hidden state $h(t) \in \mathbb{R}^{d_{\mathrm{ssm}} \times 1}$, where $d_{\mathrm{ssm}}$ denotes the SSM state dimension. Its continuous-time dynamics are defined by the following linear ordinary differential equations (ODEs):
\begin{equation}
    \begin{aligned}
        h'(t) & = \mathbf{A} h(t) +\mathbf{B} x(t), \\
        y(t)  & =\mathbf{C} h(t),
    \end{aligned}
    \label{eq:ssm}
\end{equation}
where $\mathbf{A} \in \mathbb{R}^{d_{\mathrm{ssm}} \times d_{\mathrm{ssm}}}$ is the evolution parameter. $\mathbf{B} \in \mathbb{R}^{d_{\mathrm{ssm}} \times 1}$ and $\mathbf{C} \in \mathbb{R}^{1 \times d_{\mathrm{ssm}}}$ are the projection parameters. The symbol $d_{\mathrm{ssm}}$ is distinct from the encoder token width $d$ used below.

For use in deep neural networks, the continuous system is discretized with a zero-order hold. Specifically, the continuous parameters $\mathbf{A}$ and $\mathbf{B}$ are converted to discrete parameters $\overline{\mathbf{A}}$ and $\overline{\mathbf{B}}$ using a timescale parameter $\Delta \in \mathbb{R}$:
\begin{equation}
    \begin{aligned}
        \overline{\mathbf{A}} & = \exp (\Delta\mathbf{A}),                                                         \\
        \overline{\mathbf{B}} & = (\Delta\mathbf{A})^{-1}(\exp(\Delta\mathbf{A}) - \mathbf{I}) \cdot \Delta \mathbf{B}.
    \end{aligned}
\end{equation}
The discretized form of \cref{eq:ssm} can then be written as:
\begin{equation}
    \begin{aligned}
        h_{t} & = \overline{\mathbf{A}} h_{t-1} + \overline{\mathbf{B}} x_t, \\
        y_t   & = \mathbf{C} h_t.
    \end{aligned}
\end{equation}

When $\Delta$, $\mathbf{A}$, $\mathbf{B}$, and $\mathbf{C}$ are shared across all time steps, this LTI recurrence can equivalently be evaluated using the structured convolution kernel $\overline{\mathbf{K}}$:
\begin{equation}
    \begin{aligned}
        \overline{\mathbf{K}} & = (\mathbf{C}\overline{\mathbf{B}}, \mathbf{C}\overline{\mathbf{A}}\overline{\mathbf{B}}, \dots, \mathbf{C}\overline{\mathbf{A}}^{k}\overline{\mathbf{B}}), \\
        \mathbf{y}            & = \mathbf{x} \ast  \overline{\mathbf{K}},
    \end{aligned}
\end{equation}
where $k \in [0, S-1]$ and $S$ is the length of the input sequence $\mathbf{x}$. We also have $\mathbf{y} =\left \{y_1, \dots, y_S\right \}$ and $\mathbf{x} =\left \{x_1, \dots, x_S\right \}$, while $\overline{\mathbf{K}} \in \mathbb{R}^S$ is a structured convolutional kernel.

Mamba departs from this LTI formulation by introducing a selective SSM~\cite{Mamba}. In particular, the step size and the input and output projections are conditioned on the current input:
\begin{equation}
    \Delta_t = s_{\Delta}(x_t), \qquad
    \mathbf{B}_t = s_{B}(x_t), \qquad
    \mathbf{C}_t = s_{C}(x_t),
\end{equation}
where $s_{\Delta}$, $s_B$, and $s_C$ are learned projections. Consequently, the discretized parameters become time-varying:
\begin{equation}
    \begin{aligned}
        \overline{\mathbf{A}}_t & = \exp(\Delta_t\mathbf{A}), \\
        \overline{\mathbf{B}}_t & = (\Delta_t\mathbf{A})^{-1}
        (\exp(\Delta_t\mathbf{A})-\mathbf{I}) \cdot \Delta_t\mathbf{B}_t, \\
        h_t & = \overline{\mathbf{A}}_t h_{t-1} + \overline{\mathbf{B}}_t x_t, \\
        y_t & = \mathbf{C}_t h_t.
    \end{aligned}
\end{equation}
Because these parameters vary with the input and the sequence position, Mamba cannot be evaluated using a single input-independent convolution kernel. Instead, it computes the time-varying recurrence with a hardware-aware selective-scan algorithm, preserving linear complexity in the sequence length~\cite{Mamba}.

In this paper, we use MambaMLP~\cite{ARM} as the backbone, as shown in~\cref{fig:star}. The MambaMLP block uses Mamba as the token mixer and a multilayer perceptron (MLP) as the channel mixer. During autoregressive pretraining, each block uses a single forward selective scan. This scan preserves causal information flow, making the encoder compatible with next-cluster prediction. For downstream fine-tuning, each block uses four scans to enable bidirectional global modeling, following VMamba~\cite{VMamba}.

\subsection{Autoregressive pretraining}
We first review standard autoregressive pretraining and its adaptation to visual data, and then introduce STAR.

\subsubsection{Next Token Prediction}
Next-token prediction is the standard objective in autoregressive (AR) pretraining and underlies many large language models (LLMs)~\cite{gpt4,llama2}. Each token is predicted from the preceding tokens. For a sequence $\mathcal{U}=\{u_1,\dots,u_n\}$, the joint probability factorizes into conditional probabilities:
\begin{equation}
    \Pr(u) = \prod_{k=1}^{n}\Pr(u_k\mid u_{< k}, f_{\theta}),
\end{equation}
where $u_{< k}$ denotes the set of the first $k-1$ tokens and is the context used to predict the $k^{\mathrm{th} }$ token. $f_{\theta}$ is a neural network with parameters $\theta$.

Autoregressive pretraining computes the likelihood of each token $u_k$ based on the context of all preceding tokens from $u_1$ to $u_{k-1}$ and minimizes the negative log-likelihood:
$$
    \mathcal{L}  =-\log{\Pr(u)}.
$$

\subsubsection{Autoregressive pretraining in Vision}
Images are inherently two-dimensional and lack a unique sequential order. Visual autoregressive pretraining therefore requires a serialization rule. Prior methods, including iGPT, AIM, and ARM, operate at different levels of sequence granularity. We adopt the cluster-level formulation of ARM as our starting point.

The main notation used in our visual autoregressive formulation and STAR is summarized in \cref{tab:notation}.

\begin{table}[t!]
    \centering
    \caption{Main notation used in visual autoregressive pretraining and STAR.}
    \label{tab:notation}
    \footnotesize
    \setlength{\tabcolsep}{4pt}
    \renewcommand{\arraystretch}{0.92}
    \begin{tabular}{p{0.16\linewidth}p{0.51\linewidth}p{0.23\linewidth}}
        \toprule
        Symbol & Description & Dimension or relation \\
        \midrule
        $\mathcal{D}_u$ & Unlabeled pretraining dataset. & -- \\
        $X$ & An image sampled from $\mathcal{D}_u$. & -- \\
        $C$ & Number of image channels. & $C=3$ \\
        $p$ & Patch size in pixels. & Default: $p=16$ \\
        $n$ & Number of patches in an image. & $n\in\mathbb{N}^{+}$ \\
        $P_i$ & The $i$-th flattened image patch. & $P_i\in\mathbb{R}^{s}$ \\
        $s$ & Dimension of a flattened patch. & $s=p^{2}C$ \\
        $d$ & Width of each encoder input token. & $d=s=768$ here \\
        $e$ & Patch embedding applied independently to each flattened patch. & $e:\mathbb{R}^{s}\rightarrow\mathbb{R}^{d}$ \\
        $c_i$ & The $i$-th raw-patch content cluster of an image. & $c_i\in\mathbb{R}^{m\times s}$ \\
        $\widehat c_i$ & Embedded content cluster after patch embedding and positional encoding. & $\widehat c_i\in\mathbb{R}^{m\times d}$ \\
        $H_c,W_c$ & Spatial height and width of a cluster in pixels. & Default: $H_c=W_c=4p$ \\
        $m$ & Number of patches in each cluster. & $m=\frac{H_c}{p}\frac{W_c}{p}$ \\
        $\widehat c_{\leq i}$ & Embedded ARM prefix through the $i$-th content cluster. & $\widehat c_{\leq i}\in\mathbb{R}^{mi\times d}$ \\
        $L$ & Number of content clusters in each image. & Default: $L=9$ \\
        $f$ & MambaMLP encoder used for visual autoregressive pretraining. & $\forall\ell:\ \mathbb{R}^{m\ell\times d}\rightarrow\mathbb{R}^{m\ell\times d}$ \\
        $g$ & Autoregressive decoder that predicts the next target cluster from encoder features. & $\forall\ell:\ \mathbb{R}^{m\ell\times d}\rightarrow\mathbb{R}^{m\times s}$ \\
        $\mathcal{L}$ & Single-image visual autoregressive pretraining loss. & Scalar \\
        $c_{\mathrm{sp}}$ & Shared separator cluster; it is both an encoder input and a reconstruction target because $d=s$. & $c_{\mathrm{sp}}\in\mathbb{R}^{m\times d}=\mathbb{R}^{m\times s}$ \\
        $c_0^j$ & Encoder-input separator prepended to image $j$. & $c_0^j:=c_{\mathrm{sp}}\in\mathbb{R}^{m\times d}$ \\
        $\mathbf{0},\mathbf{1}$ & All-zeros and all-ones vectors used to construct a separator cluster. & $\mathbf{0},\mathbf{1}\in\mathbb{R}^{d}=\mathbb{R}^{s}$ \\
        $X_j$ & Raw content-cluster sequence of image $j$. & $X_j=\{c_1^j,\dots,c_L^j\}$ \\
        $N$ & Number of images in an input sequence. & $N\in\mathbb{N}^{+}$ \\
        $\mathcal{I}_{\mathrm{enc}}$ & Concatenated multi-image tensor input to the encoder. & $\mathcal{I}_{\mathrm{enc}}\in\mathbb{R}^{mT\times d}$ \\
        $z_i$ & The $i$-th cluster unit in $\mathcal{I}_{\mathrm{enc}}$. & $z_i\in\mathbb{R}^{m\times d}$ \\
        $\tau_{i+1}$ & Reconstruction target after encoder prefix $z_{\le i}$. & $\tau_{i+1}\in\mathbb{R}^{m\times s}$ \\
        $T$ & Cluster-unit sequence length. & $T=N(L+1)$ \\
        $\tau_{T+1}$ & Additional terminal separator used only as the final reconstruction target. & $\tau_{T+1}:=c_{\mathrm{sp}}\in\mathbb{R}^{m\times s}$ \\
        $\mathcal{L}_{\mathrm{STAR}}$ & STAR long-sequence autoregressive pretraining loss. & Scalar \\
        \bottomrule
    \end{tabular}
\end{table}

Given a natural image $X$ from an unlabeled dataset $\mathcal{D} _u$, we first reshape it into $n$ patches, denoted as $X=\left \{P_1,\dots,P_n\right \} \in \mathbb{R}^{n\times s}$, where $s=p^2C$ is the dimension of a flattened patch and $P_i \in \mathbb{R}^{s}$ is the $i$-th patch. The patch embedding $e:\mathbb{R}^{s}\rightarrow\mathbb{R}^{d}$ is applied independently to every patch. To encapsulate the 2D spatial information at the token level, we group spatially adjacent patches into content clusters that serve as the prediction units, as illustrated in \cref{fig:token}. Let $c_i\in\mathbb{R}^{m\times s}$ be the $i$-th raw-patch cluster, where $[c_i]_a\in\mathbb{R}^{s}$ is its $a$-th patch and each cluster contains $m$ patches. Let $\widehat c_i\in\mathbb{R}^{m\times d}$ denote $c_i$ after applying $e$ row-wise and adding its positional encoding, and define the embedded prefix $\widehat c_{\le i}:=\operatorname{concat}(\widehat c_1,\dots,\widehat c_i)\in\mathbb{R}^{mi\times d}$. Finally, define $\operatorname{MSE}(u,v):=\frac{1}{s}\lVert u-v\rVert_2^2$ for $u,v\in\mathbb{R}^{s}$. For the clustered raw-patch sequence $X=\left \{c_1,\dots,c_L\right \}$, the ARM loss is
\begin{equation}
    \begin{aligned}
        \mathcal{L}
        = \frac{1}{L-1}\sum_{i=1}^{L-1}\frac{1}{m}\sum_{a=1}^{m}
        \operatorname{MSE}\!\left(
        \left[g\!\left(f(\widehat c_{\leq i})\right)\right]_a,
        [c_{i+1}]_a\right),
    \end{aligned}
\end{equation}
where $m=\frac{H_c}{p}\frac{W_c}{p}$, with $p=16$ and $H_c=W_c=4p$ in our configuration. At prediction step $i$, the encoder maps $\widehat c_{\le i}\in\mathbb{R}^{mi\times d}$ to features in $\mathbb{R}^{mi\times d}$, and the decoder maps these prefix features to a next-cluster prediction in $\mathbb{R}^{m\times s}$, matching the target $c_{i+1}\in\mathbb{R}^{m\times s}$. Thus, the reconstruction error is averaged over the $s$ dimensions of each patch, the $m$ patches within each target cluster, and the $L-1$ valid next-cluster targets.

\subsection{Separator for AR}
First, in vanilla ARM, the first cluster $c_1$ is used only as context for predicting subsequent clusters and is not itself a prediction target. Consequently, no reconstruction loss is computed for $c_1$. The final content cluster $c_L$ is predicted from its preceding context but has no successor, so it does not produce an outgoing next-cluster loss. Second, the fixed image size produces a relatively short encoder sequence for each image and does not fully exploit the ability of Mamba models to process long sequences. We introduce separators to address these two limitations.

For each raw content-cluster sequence $X_j=\left\{c_1^j,\dots,c_L^j \right\}$, we add a \textit{separator cluster} $c_{\mathrm{sp}}\in\mathbb{R}^{m\times d}$ that is created directly in the encoder input space, where $d$ is the encoder token width. It therefore occupies the same number of positions as an embedded content cluster $\widehat c_i^j\in\mathbb{R}^{m\times d}$ and is concatenated with the embedded content clusters without passing through the patch embedding $e$ or any additional projection. In the final STAR configuration, $c_{\mathrm{sp}}$ is the Identity separator: its diagonal positions are all-ones vectors, and all remaining positions are all-zeros vectors. The Identity separator uses positional encoding, and all tokens within it share the same positional encoding. All images share exactly the same separator, i.e., $c_0^j:=c_{\mathrm{sp}}$ for every $j\in[1,N]$. The resulting encoder-input sequence for image $j$ is $\widehat X_j:=\operatorname{concat}(c_0^j,\widehat c_1^j,\dots,\widehat c_L^j)\in\mathbb{R}^{m(L+1)\times d}$.

With the addition of $c_{\mathrm{sp}}$, we can represent multiple images within the same sequence, using $c_{\mathrm{sp}}$ as a separator. During pretraining, no other special token is inserted between images. The patch positional encodings are reset at each image boundary, so every image reuses the same positional-encoding pattern. We concatenate the per-image encoder sequences to obtain
\[
    \mathcal{I}_{\mathrm{enc}}
    :=\operatorname{concat}(\widehat X_1,\dots,\widehat X_N)
    \in\mathbb{R}^{mT\times d},
    \qquad T=N(L+1).
\]
At the cluster-unit level, write $\mathcal{I}_{\mathrm{enc}}=\operatorname{concat}(z_1,\dots,z_T)$, where $z_i\in\mathbb{R}^{m\times d}$ and $z_{\le i}:=\operatorname{concat}(z_1,\dots,z_i)\in\mathbb{R}^{mi\times d}$. For every prefix $z_{\le i}$, define the next-cluster reconstruction target $\tau_{i+1}\in\mathbb{R}^{m\times s}$. If the next cluster unit is content cluster $c_k^j$, then $\tau_{i+1}$ contains its normalized raw patches; if it is a separator, then $\tau_{i+1}=c_{\mathrm{sp}}$. Because $d=s$ in our configuration, the same predefined separator values can be used in both the encoder-input and reconstruction-target spaces. We additionally define $\tau_{T+1}:=c_{\mathrm{sp}}\in\mathbb{R}^{m\times s}$ as a terminal separator target; it is not part of $\mathcal{I}_{\mathrm{enc}}$ and is not counted in $T$. Therefore, the STAR loss is
\begin{equation}
    \mathcal{L}_{\mathrm{STAR}}
    = \frac{1}{T}\sum_{i=1}^{T}
    \frac{1}{m}\sum_{a=1}^{m}
    \operatorname{MSE}\!\left(
    \left[g\!\left(f(z_{\leq i})\right)\right]_a,
    [\tau_{i+1}]_a\right),
\end{equation}
Here $T=N(L+1)$ is the number of encoder-input cluster units, including one separator cluster per image. The first unit $z_1=c_0^1$ is used only as the initial context. For each non-final image, its last content cluster predicts the separator prepended to the subsequent image, giving $N-1$ inter-image separator targets. The last content cluster of the final image predicts the terminal target $\tau_{T+1}=c_{\mathrm{sp}}$. Consequently, all $N$ separator prediction errors are included in $\mathcal{L}_{\mathrm{STAR}}$. At step $i$, $f$ maps $z_{\le i}\in\mathbb{R}^{mi\times d}$ to features of the same shape, and $g$ maps those features to a prediction in $\mathbb{R}^{m\times s}$, matching $\tau_{i+1}$. The factor $1/T$ averages over all $T$ valid next-cluster targets. Following MAE~\cite{MAE}, we independently normalize the pixel values of each image patch within a content-cluster target before computing $\mathcal{L}$ and $\mathcal{L}_{\mathrm{STAR}}$; separator targets are not normalized. Since $s=p^{2}C=d=768$ in our configuration, the same predefined tensor $c_{\mathrm{sp}}$ serves as both encoder input and reconstruction target, so no projection is required on either side.

The complete framework is illustrated in~\cref{fig:star}. Unlike vanilla ARM, STAR provides an outgoing prediction target for the final content cluster of every image, including the final image in the packed sequence. Packing multiple images also increases the context length without changing the resolution of any individual image. We evaluate this design on classification, robustness, and dense prediction.

\section{Experiment}
\label{sec:exp}

\subsection{Experiment Settings}
\paragraph{Pretraining}
We conduct autoregressive pretraining on the ImageNet-1K training set~\cite{ImageNet}, which consists of approximately 1.28 million images. Similar to the approach used in MAE, we apply two data augmentation techniques, namely random cropping and horizontal flipping. Throughout all experiments, we utilize MambaMLP as the backbone. During the autoregressive pretraining phase, the MambaMLP blocks are configured to include only a single-scan Mamba layer, aligning with the unidirectional modeling approach. We employ an AdamW optimizer with a cosine learning-rate scheduler and train for 200 epochs unless otherwise specified. The global batch size is fixed at 2,048 images. For an input sequence containing $N$ images, the corresponding number of multi-image sequences per batch is $2048/N$. Therefore, varying $N$ does not change the number of training images or optimization steps in each epoch. The initial learning rate is set to $1.5\times10^{-4}\times\frac{\mathrm{batch\ size}}{256}$, with a weight decay of 0.05, $\beta_1=0.9$, and $\beta_2=0.95$. We use a 40-epoch warm-up. During autoregressive pretraining, images are randomly cropped to $192\times192$. With a patch size of $16\times16$, each image is converted into a $12\times12$ grid containing 144 patch tokens.

The final STAR configuration uses a separator cluster with $4 \times 4$ patches and an input sequence containing 4 images. Accordingly, each batch contains 512 multi-image sequences and 2,048 images in total. The separator-value ablation in \cref{tab:value}, the separator-position ablation in \cref{tab:pos}, and the pretraining-duration comparison in \cref{tab:duration} use 8 images per input sequence, so that each batch holds 256 multi-image sequences and again 2,048 images in total. The number-of-images ablation in \cref{tab:num} evaluates sequences containing 1 to 16 images. The per-setting configurations are collected in \cref{tab:config}. Our decoder is a lightweight, 4-layer network based on cross-attention mechanisms, where each layer has a width of 512.

\paragraph{Fine-tuning} During downstream classification fine-tuning, images are cropped to $224\times224$, producing a $14\times14$ grid of patch tokens with a patch size of $16\times16$. We use fixed two-dimensional sine--cosine positional embeddings. When transferring the model from $192\times192$ autoregressive pretraining to $224\times224$ fine-tuning, we reshape the pretrained patch positional embeddings from their $12\times12$ grid and resize them to the $14\times14$ fine-tuning grid using two-dimensional bicubic interpolation. The resulting positional embeddings remain fixed during fine-tuning. Cluster grouping and separator clusters are used only during autoregressive pretraining to organize the multi-image sequence and define cluster-level prediction targets. During downstream classification fine-tuning, patch tokens are not grouped into clusters, and no separator clusters are inserted; the encoder directly processes the standard $14\times14$ patch-token grid. The classification representation is obtained by mean-pooling all patch tokens before the linear classification head. During fine-tuning, we use AdamW for 100 epochs with a cosine learning-rate scheduler and a 5-epoch warm-up. The hyperparameters are a global batch size of 1,024, an initial learning rate of $5\times10^{-4}\times\frac{\mathrm{batch\ size}}{256}$, a weight decay of 0.05, a stochastic-depth rate of 0.1, and a layer-wise learning-rate decay of 0.65. We follow the data augmentations used in MAE, including RandAug, Mixup, CutMix, and label smoothing, and use an exponential moving average (EMA) of the model parameters. To adapt Mamba to two-dimensional visual data during downstream fine-tuning, we follow VMamba and use four scanning paths in MambaMLP.

\paragraph{COCO transfer evaluation} We additionally evaluate whether the pretrained representation transfers beyond ImageNet-1K classification. We evaluate object detection and instance segmentation on COCO 2017~\cite{coco}. Following the experimental setup of ARM~\cite{ARM}, we integrate ARM or STAR as the backbone of Mask R-CNN~\cite{maskrcnn} and adopt the downstream protocol of Swin Transformer~\cite{swint}. For both backbones, Mask R-CNN is initialized from the corresponding checkpoint obtained after 1,600 epochs of autoregressive pretraining, namely the 1,600-epoch ARM and STAR entries in~\cref{tab:ssl}. We fine-tune every model for 36 epochs (the $3\times$ schedule) with multi-scale training, randomly resizing the shorter image side between 480 and 800 pixels. Optimization uses AdamW with an initial learning rate of $1\times10^{-4}$, a weight decay of 0.05, and a total batch size of 16. All augmentations and downstream hyperparameters are kept identical across backbones to isolate the effect of autoregressive pretraining. We report the standard box AP ($\mathrm{AP}^{\mathrm{bbox}}$) and mask AP ($\mathrm{AP}^{\mathrm{mask}}$) metrics.

\paragraph{Out-of-distribution evaluation} The OOD experiments use the ImageNet-1K fine-tuned ARM-B and STAR-B models initialized from the corresponding 1,600-epoch ARM and STAR checkpoints in~\cref{tab:ssl}. We directly evaluate these fixed fine-tuned models on ImageNet-V2~\cite{ImageNetV2}, ImageNet-A~\cite{ImageNetA}, ImageNet-R~\cite{ImageNetR}, ImageNet-Sketch~\cite{ImageNetSketch}, and ImageNet-ReaL~\cite{ImageNetReaL} and report top-1 accuracy. No additional fine-tuning, adaptation, or target-specific hyperparameter selection is performed on any of these variants; thus, the results measure how well the ImageNet-1K fine-tuned representations generalize beyond the original validation distribution.

\paragraph{Unified configurations} \cref{tab:config} summarizes the configurations used in all our experiments. Every autoregressive pretraining run uses $192\times192$ inputs, a patch size of $16\times16$, and content clusters of $4\times4$ patches, so each image contributes a $12\times12$ grid of 144 content patch tokens grouped into $L=9$ content clusters, and each separator cluster contributes 16 separator patch tokens. Under the SC layout, the patch-token sequence length for $N$ images is therefore $160N$. Because the global batch size is fixed at 2,048 images and each batch correspondingly holds $2048/N$ multi-image sequences, all configurations in \cref{tab:config} process the same number of images per optimization step and the same number of optimization steps per epoch; only the patch-token sequence length seen by the encoder changes with $N$. The final STAR configuration is the Identity separator cluster of $4\times4$ patches placed at the start of each image with $N=4$ images per input sequence, and it is the configuration reported for STAR in \cref{tab:ssl,tab:super,tab:coco,tab:ood}. The separator studies instead use their own defaults: \cref{tab:type,tab:num} vary the separator form and $N$ with the simplest \textit{Zeros} separator, while \cref{tab:value,tab:pos,tab:duration} fix $N=8$ so that separator values, positions, and pretraining durations are compared at an identical patch-token sequence length.

\begin{table}[t!]
    \centering
    \caption{\textbf{Unified experimental configurations.} Resolution and patch size are given in pixels, and cluster size in patches. Patch-token length counts patch-token positions, including separator patch tokens, and is not a count of cluster prediction units. Cluster grouping and separator clusters are used only during autoregressive pretraining. For the separator-type ablation, 576, 580, and 640 denote the \textit{None}, \textit{Token}, and \textit{Cluster} variants, respectively. $^{\ddag}$1{,}280 corresponds to the SC and CS layouts, which insert one separator cluster per image; the denser SCS and CSC layouts insert one separator cluster every $r=3$ content clusters, giving 192 patch tokens per image and 1{,}536 per sequence.}
    \label{tab:config}
    \resizebox{1\linewidth}{!}{
        \begin{tabular}{l|c|c|c|c|c}
            \toprule
            Setting                                       & Resolution & Patch size & Cluster size & Images/sequence & Patch-token length \\
            \midrule
            ARM baseline                                  & 192        & 16         & $4\times4$   & 1               & 144                \\
            Separator-type ablation (\cref{tab:type})     & 192        & 16         & $4\times4$   & 4               & 576 / 580 / 640    \\
            Separator-value ablation (\cref{tab:value})   & 192        & 16         & $4\times4$   & 8               & 1{,}280            \\
            Separator-position ablation (\cref{tab:pos})  & 192        & 16         & $4\times4$   & 8               & 1{,}280$^{\ddag}$  \\
            Separator-number ablation (\cref{tab:num})    & 192        & 16         & $4\times4$   & 1--16           & 160--2{,}560       \\
            Pretraining-duration comparison (\cref{tab:duration}) & 192 & 16 & $4\times4$ & 8 & 1{,}280 \\
            Final STAR (\cref{tab:ssl,tab:super,tab:coco,tab:ood}) & 192 & 16 & $4\times4$ & 4 & 640 \\
            Downstream fine-tuning                        & 224        & 16         & N/A          & 1               & 196                \\
            \bottomrule
        \end{tabular}}
\end{table}

\subsection{Main Properties}
Because cluster-level separators for multi-image autoregressive pretraining with Vision Mamba remain underexplored, we study several key design aspects under our available computational budget, including separator type, value, position, and number.

\paragraph{Separator type} We first evaluate how the separator type affects representation learning. \cref{tab:type} reports the fine-tuning and EMA accuracies for different separator types. In these experiments, we concatenate four images into a single sequence and compare no separator, token-level separators, and cluster-level separators.

\begin{table}[t!]
    \centering
    \caption{\textbf{Fine-tuning accuracy (\%) and EMA accuracy (\%) of pretrained models by different types of separators on ImageNet-1K.}
        The ARM baseline uses one image with 144 patch tokens. The \textit{None}, \textit{Token}, and \textit{Cluster} variants use 4 images with patch-token sequence lengths of 576, 580, and 640, respectively. The best result in each column is \textbf{bolded}.}
    \label{tab:type}
    \resizebox{0.92\linewidth}{!}{
        \begin{tabular}{l|c|c|c|c}
            \toprule
            Type     & Backbone   & Epochs & FT (\%)        & EMA (\%)       \\
            \midrule
            Baseline & MambaMLP-B & 200    & 81.81          & 82.15          \\
            None     & MambaMLP-B & 200    & 78.64          & 78.63          \\
            Token    & MambaMLP-B & 200    & 79.41          & 79.37          \\
            Cluster  & MambaMLP-B & 200    & \textbf{82.15} & \textbf{82.40} \\
            \bottomrule
        \end{tabular}}
\end{table}

We first examine the effect of extending the input sequence without adding a separator. Concatenating four image sequences decreases EMA accuracy from $82.15\%$ to $78.63\%$, showing that naive multi-image concatenation is ineffective in this setting. We then introduce an input-space zero-vector token as the separator. EMA accuracy increases to $79.37\%$, an improvement of 0.74 percentage points over the configuration without separators. This result indicates that explicitly marking image boundaries benefits long-sequence autoregressive modeling. However, the token-based separator still underperforms single-image autoregressive training. We hypothesize that this gap arises from a mismatch between the single-token separator and the $4\times4$-token cluster prediction unit.

Finally, we set the separator as a cluster. An input-space zero-vector token is expanded into a $4\times4$ input-space separator cluster. Consistent with all separator variants, the tokens within this cluster share the same positional encoding. We place the cluster-based separator at the start of each image, as illustrated in~\cref{fig:star}, and reconstruct each cluster sequentially. The cluster-based separator achieves an accuracy of 82.15\% and an EMA accuracy of 82.40\%, representing improvements of 0.34\% and 0.25\% over autoregressive pretraining on a single image, respectively. Among the evaluated designs, the cluster-based separator configuration performs best. Because the \textit{None}, \textit{Token}, and \textit{Cluster} variants have different sequence lengths, this comparison reflects their complete configurations rather than separator granularity alone.

\begin{figure}[t!]
    \centering
    \includegraphics[width=0.8\linewidth]{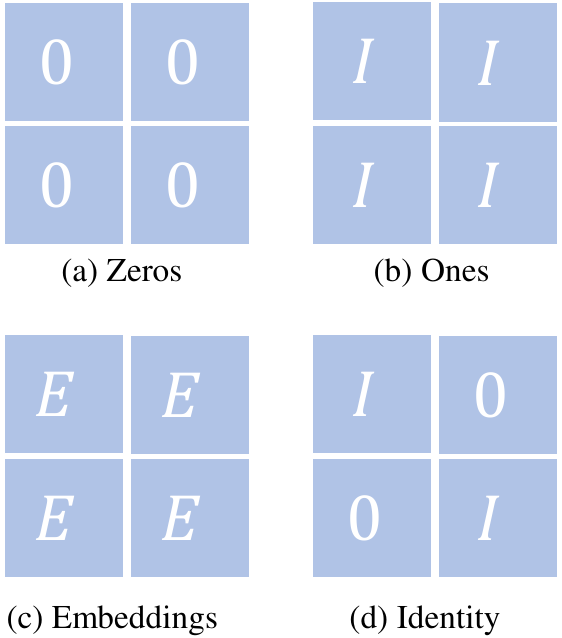}
    \caption{\textbf{Input-space definitions of different separator values.} \textbf{Zeros}: every position is a zero vector. \textbf{Ones}: every position is a one vector. \textbf{Embeddings}: every position uses the same learnable \textit{nn.Embedding(1, d)} vector. \textbf{Identity}: the diagonal positions are all-ones vectors, and all remaining positions are all-zeros vectors.}
    \label{fig:value}
\end{figure}

\paragraph{Separator value} The appropriate value of a cluster-level separator in multi-image Vision Mamba autoregressive pretraining has not been established. A separator should remain distinct from content representations while clearly marking image boundaries. All four separator variants are created directly in the encoder input space and occupy the same number of positions as a content cluster: \textit{Zeros} assigns a zero vector to every cluster position; \textit{Ones} assigns a one vector to every position; \textit{Embeddings} assigns the same learnable \textit{nn.Embedding(1, d)} vector to every position; and \textit{Identity} assigns all-ones vectors to the diagonal positions and all-zeros vectors to all remaining positions. All four variants use positional encoding, with the same positional encoding shared across all tokens within each separator. \cref{fig:value} illustrates these four input-space separator definitions.

\begin{table}[t!]
    \centering
    \caption{\textbf{Fine-tuning accuracy (\%) and EMA accuracy (\%) of pretrained models by different values of separators on ImageNet-1K.}
        All separator variants use 8 images per input sequence. The best result in each column is \textbf{bolded}. The second-best result in each column is \underline{underlined}.}
    \label{tab:value}
    \resizebox{1\linewidth}{!}{
        \begin{tabular}{l|c|c|c|c}
            \toprule
            Value      & Backbone   & Epochs & FT (\%)           & EMA (\%)          \\
            \midrule
            Baseline   & MambaMLP-B & 200    & 81.81             & 82.15             \\
            \midrule
            Zeros      & MambaMLP-B & 200    & 81.77             & \textbf{82.35}    \\
            Ones       & MambaMLP-B & 200    & 81.83             & 82.26             \\
            Embeddings & MambaMLP-B & 200    & \textbf{82.05}    & 82.27             \\
            Identity   & MambaMLP-B & 200    & \underline{82.03} & \underline{82.32} \\
            \midrule
            Zeros      & MambaMLP-B & 300    & 81.79             & 82.54             \\
            Identity   & MambaMLP-B & 300    & \textbf{82.08}    & \textbf{82.58}    \\
            \bottomrule
        \end{tabular}}

\end{table}

We concatenate 8 images into a single long sequence. All these separators are created directly at the encoder input width $d$, are placed at the beginning of each image, and therefore bypass the patch embedding layer. As shown in~\cref{tab:value}, all four separator values improve EMA accuracy over the baseline, whereas the fine-tuning results are mixed: \textit{Zeros} and \textit{Ones} remain below the baseline, while \textit{Embeddings} and \textit{Identity} exceed it. Specifically, the \textit{Zeros} separator achieves the highest EMA accuracy at $82.35\%$, while the \textit{Embeddings} separator achieves the highest fine-tuning accuracy at $82.05\%$. The \textit{Identity} separator ranks second in both fine-tuning accuracy and EMA accuracy.

We further compare the separator values after extending pretraining to 300 epochs. The \textit{Identity} separator provides the best overall performance, with fine-tuning and EMA accuracies of $82.08\%$ and $82.58\%$, respectively. Considering both metrics, we use the \textit{Identity} separator in the final STAR configuration. For the remaining separator ablations, we retain the simpler \textit{Zeros} separator as the study default.

\begin{figure}[t!]
    \centering
    \caption{\textbf{Different positions of the separator cluster.} \textbf{SC}: a single separator cluster is placed before all $L$ content clusters of an image. \textbf{CS}: a single separator cluster is placed after all $L$ content clusters of an image. \textbf{SCS} and \textbf{CSC} are denser layouts that insert one separator cluster for every group of $r=3$ content clusters, giving three separator clusters per image: \textbf{SCS} places each separator before its group of content clusters, whereas \textbf{CSC} places each separator after its group. SC and CS therefore use 160 patch tokens per image, while SCS and CSC use 192.}
    \label{fig:pos}
    \includegraphics[width=0.95\linewidth]{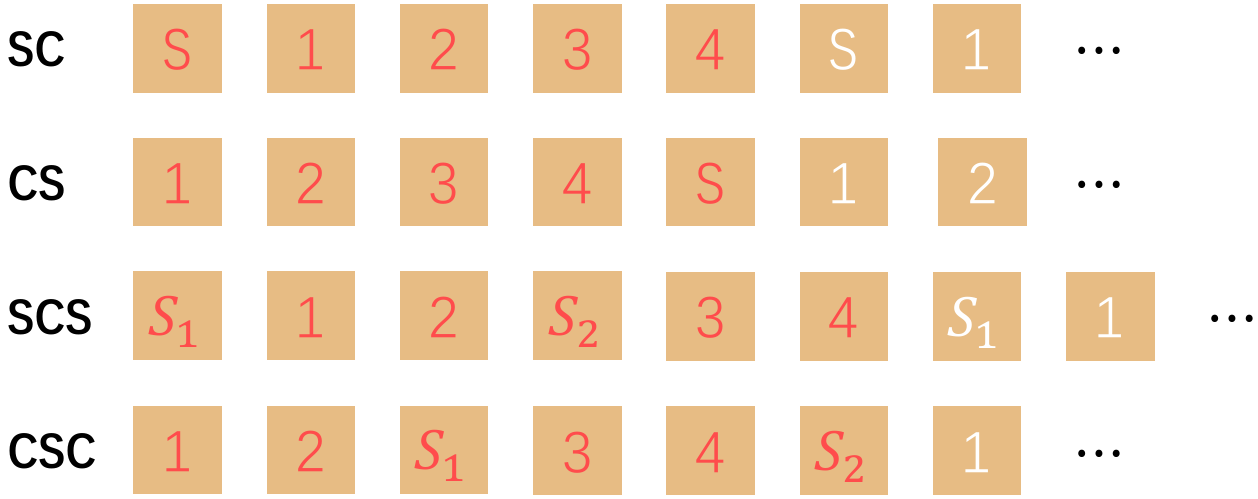}

\end{figure}

\paragraph{Separator position} We conduct an in-depth investigation into the insertion positions of the separator. All position variants concatenate 8 images into a single sequence and use the \textit{Zeros} separator. Two primary configurations are considered: placing the separator before the other clusters within an image, termed as \textbf{SC}, and placing it after the clusters, referred to as \textbf{CS}. Additionally, we explore denser separator layouts. Let $r=\sqrt{L}$ and partition the $L$ serialized content clusters into $r$ consecutive groups of $r$ clusters each. In our setting, $L=9$ and $r=3$. We insert one separator cluster for every group, giving three separator clusters per image. This dense configuration is further divided into two variations: placing each separator before its group of content clusters, denoted as \textbf{SCS}, and after its group, denoted as \textbf{CSC}. \cref{fig:pos} illustrates these different layout strategies in detail.

\begin{table}[t!]
    \centering
    \caption{\textbf{Fine-tuning accuracy (\%) and EMA accuracy (\%) of pretrained models by different positions of separators on ImageNet-1K.}
        All position variants use 8 images per input sequence. The best result in each column is \textbf{bolded}. The second-best result in each column is \underline{underlined}.}
    \label{tab:pos}
    \resizebox{0.92\linewidth}{!}{
        \begin{tabular}{l|c|c|c|c}
            \toprule
            Position & Epochs & Time (h) & FT (\%)           & EMA (\%)          \\
            \midrule
            SC       & 200    & 24.6     & \underline{81.77} & \textbf{82.35 }   \\
            CS       & 200    & 24.6     & 77.93             & 77.94             \\
            SCS      & 200    & 31.4     & \textbf{81.98 }   & \underline{82.33} \\
            CSC      & 200    & 31.4     & 78.69             & 78.64             \\
            \bottomrule
        \end{tabular}}

\end{table}

\cref{tab:pos} reports fine-tuning accuracy (FT), exponential-moving-average accuracy (EMA), and training time on ImageNet-1K for different separator positions. The SC configuration obtains the highest EMA accuracy (82.35\%) and the second-highest FT accuracy (81.77\%) after 24.6 hours of training. SCS obtains the highest FT accuracy (81.98\%) and the second-highest EMA accuracy but requires 31.4 hours. CS and CSC yield accuracies below 79\%. These results show that separator position substantially affects performance and that SC provides the best trade-off between accuracy and training time among the evaluated configurations. One possible explanation is that placing the separator before the content clusters preserves the complete reconstruction target for each image. When it is placed after the clusters, the first cluster of the first image in the sequence is excluded from prediction.
% We explore this further in~\cref{tab:pos}. Results demonstrate that even under the SC configuration, excluding the first cluster of all images from loss computation leads to a noticeable drop in performance. 

\paragraph{Separator number} To examine the effect of sequence length, we vary the number of images per sequence from 1 to 16. Under the SC cluster-separator layout, each image contributes 144 content patch tokens and 16 separator patch tokens, giving a total patch-token count of 160 per image. These counts refer to patch-token positions rather than cluster prediction units.

\begin{table}[t!]
    \centering
    \caption{\textbf{Fine-tuning and EMA top-1 accuracies (\%) for different numbers of images per input sequence on ImageNet-1K.}
        Num denotes the number of images per input sequence, and all rows use the cluster-based \textit{Zeros} separator in the SC layout. The best result in each column is \textbf{bolded}.}
    \label{tab:num}
    \resizebox{0.92\linewidth}{!}{
        \begin{tabular}{l|c|c|c|c}
            \toprule
            Num & Backbone   & Epochs & FT (\%)        & EMA (\%)       \\
            \midrule
            Baseline & MambaMLP-B & 200    & 81.81          & 82.15          \\
            1   & MambaMLP-B & 200    & 81.94          & 82.20          \\
            2   & MambaMLP-B & 200    & 81.92          & 82.24          \\
            4   & MambaMLP-B & 200    & \textbf{82.15} & \textbf{82.40} \\
            8   & MambaMLP-B & 200    & 81.77          & 82.35          \\
            16  & MambaMLP-B & 200    & 81.55          & 81.97          \\
            \bottomrule
        \end{tabular}}

\end{table}

Consequently, the patch-token sequence length in our experiments ranges from 160 tokens for a single image to 2,560 tokens for 16 images. \cref{tab:num} shows how performance varies with sequence length. With four images and 640 tokens, the fine-tuning and EMA accuracies reach their highest values among the evaluated settings, at $82.15\%$ and $82.4\%$, respectively. Increasing the sequence length further does not improve performance: at 2,560 tokens, both metrics decline. This result suggests that MambaMLP-B may be approaching its practical sequence-length limit under the present training configuration.

\paragraph{Pretraining duration} We additionally compare 200 and 300 epochs of autoregressive pretraining under the separator-study default, i.e., the \textit{Zeros} separator with 8 images per input sequence. All results in~\cref{tab:duration} use $192\times192$ inputs for autoregressive pretraining and $224\times224$ inputs for downstream classification fine-tuning. Extending pretraining from 200 to 300 epochs increases EMA top-1 accuracy from $82.35\%$ to $82.54\%$.

\begin{table}[t!]
    \centering
    \caption{\textbf{Fine-tuning accuracy (\%) and EMA accuracy (\%) under different pretraining durations on ImageNet-1K.} Both rows use the separator-study default of 8 images per input sequence with the \textit{Zeros} separator. The best result in each column is \textbf{bolded}.}
    \label{tab:duration}
    \resizebox{0.60\linewidth}{!}{
        \begin{tabular}{l|c|c|c}
            \toprule
            Model       & Epochs & FT (\%)         & EMA (\%)        \\
            \midrule
            STAR-B      & 200    & 81.77           & 82.35           \\
            STAR-B      & 300    & \textbf{81.79}  & \textbf{82.54} \\
            \bottomrule
        \end{tabular}}
\end{table}

\subsection{Comparisons with Previous Results}
\paragraph{Comparisons with self-supervised methods} We compare STAR with contrastive learning, MAE, and ARM using MambaMLP-B, and include published ViT-B results for context. During STAR pretraining, the input resolution is $192\times192$; downstream fine-tuning uses $224\times224$ inputs. All MambaMLP-B training times in~\cref{tab:ssl} were measured on eight NVIDIA A800 GPUs. At 300 epochs, STAR reaches 82.9\% EMA top-1 accuracy, 0.4 percentage points above ARM, while requiring 50 hours rather than 34 hours. At 1,600 epochs, STAR improves over ARM by 0.3 points (83.5\% versus 83.2\%) while requiring 267 hours rather than 182 hours. Thus, at matched epoch counts, STAR uses approximately $1.47\times$ the ARM pretraining time for gains of 0.4 and 0.3 points, respectively. For additional context, at 300 epochs STAR is $6.6\times$ and $1.4\times$ faster than the reproduced contrastive-learning and MAE runs. This trade-off reflects the cost of extending the patch-token sequence length from 144 in ARM to 640 in STAR without changing the per-image input resolution.

\begin{table}[H]
    \centering
    \captionsetup{font=small,skip=4pt}
    \caption{\textbf{EMA top-1 accuracy (\%) and pretraining time of representative methods on ImageNet-1K. STAR uses the final configuration with 4 images per input sequence.}
        ViT-B results are taken from the cited original sources, except that $\mathrm{AR}^{\dag}$ is reproduced from MAP~\cite{MAP}. MambaMLP-B supervised, contrastive, MAE, and ARM results are reproduced following ARM~\cite{ARM}; STAR results are ours. All MambaMLP-B times were measured on eight NVIDIA A800 GPUs. The best result in each column is \textbf{bolded}, and the second-best result is \underline{underlined}.}
    \label{tab:ssl}
    \renewcommand{\arraystretch}{0.92}
    \resizebox{0.94\linewidth}{!}{
        \begin{tabular}{l|c|l|l|c}
            \toprule
            Method               & Backbone   & Epochs & Time (h) & EMA top-1 (\%)   \\
            \midrule
            MoCo v3~\cite{mocov3} & ViT-B      & 300    & -        & {83.0}           \\
            MoCo v3~\cite{mocov3} & ViT-B      & 600    & -        & {83.2}           \\
            MAE~\cite{MAE}       & ViT-B      & 300    & -        & {82.9}           \\
            MAE~\cite{MAE}       & ViT-B      & 1{,}600 & -        & \textbf{{83.6}}  \\
            AR$^{\dag}$          & ViT-B      & 300    & -        & 82.5             \\
            \midrule
            Supervised           & MambaMLP-B & 300    & 110      & 81.2             \\
            Contrastive          & MambaMLP-B & 300    & 330      & 81.4             \\
            MAE                  & MambaMLP-B & 300    & 70       & 81.6             \\
            ARM                  & MambaMLP-B & 300    & 34       & 82.5             \\
            ARM                  & MambaMLP-B & 1{,}600 & 182      & 83.2             \\
            \midrule
            STAR                 & MambaMLP-B & 300    & 50       & {82.9}           \\
            STAR                 & MambaMLP-B & 1{,}600 & 267      & \underline{83.5} \\
            \bottomrule
        \end{tabular}}

\end{table}
For context, \cref{tab:ssl} also includes published ViT-B results. At 1,600 epochs, MAE~\cite{MAE} reports 83.6\% on ViT-B, while STAR-B reaches 83.5\% on MambaMLP-B. Because the backbones and objectives differ, this comparison is contextual rather than controlled.

\paragraph{Comparisons with other lightweight models} We compare STAR with representative convolutional, Vision Transformer, and Vision Mamba models in~\cref{tab:super}. Following the comparison protocol and reported results in ARM~\cite{ARM}, STAR-B remains competitive with these representative architectures. With 300 pretraining epochs, STAR-B matches the 82.9\% top-1 accuracy reported for RegNetY-16G~\cite{RegNetY}. Extending autoregressive pretraining to 1,600 epochs increases the STAR-B accuracy to 83.5\% on ImageNet-1K, approaching the 83.9\% reported for VMamba-B.

\begin{table}[H]
    \centering
    \captionsetup{font=small,skip=4pt}
    \caption{\textbf{Top-1 accuracy (\%) of lightweight models on ImageNet-1K.}
        STAR-B uses the final configuration with 4 images per input sequence. Results for RegNetY-16G, DeiT-B, Vim-B, MambaMLP-B, and VMamba-B are reproduced from ARM~\cite{ARM}. Throughput denotes downstream ImageNet classification inference at $224\times224$ resolution, measured on one NVIDIA A5000 GPU as reported by ARM; it is not pretraining throughput. STAR-B uses the MambaMLP-B backbone and therefore has the same downstream inference throughput. The best result in each column is \textbf{bolded}. The second-best result in each column is \underline{underlined}. $^{\dag\dag}$ indicates 1,600 pretraining epochs.}
    \label{tab:super}
    \renewcommand{\arraystretch}{0.92}
    \resizebox{0.94\linewidth}{!}{
        \begin{tabular}{l|c|c|c|c}
            \toprule
            \multirow{2}{*}{Model}     & \multirow{2}{*}{Token Mixer} & {Param.} & {Throughput} & {Top-1}          \\
                                       &                              & (M)      & (imgs/s)      & Top-1 (\%)       \\
            % \toprule
            %     Model  & Param. (M) & Throughput (imgs/s) & Top-1 (\%)\\
            \midrule
            RegNetY-16G~\cite{RegNetY} & 2D Conv.                     & 84       & 870           & {82.9}           \\
            DeiT-B~\cite{DeiT}         & Attention                    & 21       & 1073          & 81.2             \\
            Vim-B~\cite{Vim}           & Mamba                        & 98       & 890           & 81.2             \\
            MambaMLP-B~\cite{ARM}      & Mamba                        & 85       & 1301          & 81.2             \\
            VMamba-B                   & Mamba+2D Conv.               & 89       & 315           & \textbf{83.9}    \\
            \midrule
            % ARM-B & 85 & 1301 & 82.5 \\
            STAR-B                    & Mamba                        & 85       & 1301          & {82.9}           \\
            STAR-B$^{\dag\dag}$       & Mamba                        & 85       & 1301          & \underline{83.5} \\
            \bottomrule
        \end{tabular}}

\end{table}

\subsection{Transferability to COCO}
\label{sec:transfer}

We first examine whether STAR's classification gains transfer to dense visual prediction under the controlled Mask R-CNN protocol described above. Both backbones use the corresponding 1,600-epoch autoregressive-pretraining checkpoints from~\cref{tab:ssl}. As shown in~\cref{tab:coco}, STAR-B achieves 46.84 box AP and 41.45 mask AP on COCO 2017, outperforming the ARM baseline by 0.73 and 0.71 points, respectively. Because the detector, optimization, augmentation, and training schedule are shared, these gains are consistent with an improvement in the pretrained backbone. The improvements in both localization and instance delineation show that the learned representations transfer beyond image-level recognition.

\begin{center}
\begin{minipage}{0.95\linewidth}
    \captionsetup{width=\linewidth}
    \captionof{table}{\textbf{Mask R-CNN object detection and instance segmentation on COCO 2017.} Both backbones are initialized from their respective 1,600-epoch autoregressive-pretraining checkpoints in~\cref{tab:ssl} and use the same 36-epoch ($3\times$) multi-scale protocol. Results are reported using box AP ($\mathrm{AP}^{\mathrm{bbox}}$) and mask AP ($\mathrm{AP}^{\mathrm{mask}}$). The best result in each column is \textbf{bolded}.}
    \label{tab:coco}
    \centering
    \resizebox{0.58\linewidth}{!}{
        \begin{tabular}{l|cc}
            \toprule
            Method        & $\mathrm{AP}^{\mathrm{bbox}}$ & $\mathrm{AP}^{\mathrm{mask}}$ \\
            \midrule
            ARM           & 46.11                          & 40.74                          \\
            STAR-B        & \textbf{46.84}                & \textbf{41.45}                \\
            \bottomrule
        \end{tabular}}
\end{minipage}
\end{center}

\subsection{Out-of-Distribution Generalization}
\label{sec:ood}

We assess out-of-distribution generalization by directly evaluating the ImageNet-1K fine-tuned ARM-B and STAR-B models initialized from the corresponding 1,600-epoch autoregressive-pretraining checkpoints in~\cref{tab:ssl}. These fine-tuned models are kept fixed and are evaluated without any additional training or target-domain adaptation. We consider four distribution-shift benchmarks---ImageNet-V2, ImageNet-A, ImageNet-R, and ImageNet-Sketch---together with the relabeled ImageNet-ReaL benchmark. This protocol separates robustness evaluation from downstream transfer learning: unlike the COCO experiments, the OOD results require no task-specific fine-tuning.

As shown in~\cref{tab:ood}, STAR-B improves upon ARM-B on every benchmark. The gains are 0.2, 3.8, 1.5, 0.7, and 0.4 percentage points on ImageNet-V2, ImageNet-A, ImageNet-R, ImageNet-Sketch, and ImageNet-ReaL, respectively. Consequently, the unweighted mean accuracy increases by 1.7 points, from 55.1\% to 56.8\%. The largest improvements occur on ImageNet-A and ImageNet-R; this pattern is consistent with stronger gains on two substantial distribution shifts, although the benchmark set is too limited to establish a general advantage for any particular type of shift.

\begin{table}[H]
    \centering
    \captionsetup{width=0.95\linewidth}
    \caption{\textbf{Direct evaluation of ImageNet-1K fine-tuned models on ImageNet robustness benchmarks.} ARM-B and STAR-B are initialized from their respective 1,600-epoch autoregressive-pretraining checkpoints in~\cref{tab:ssl}. The resulting ImageNet-1K fine-tuned models are evaluated without additional fine-tuning or target-domain adaptation. Avg. is the unweighted mean across the five benchmarks. The best result in each column is \textbf{bolded}.}
    \label{tab:ood}
    \resizebox{1\linewidth}{!}{
        \begin{tabular}{l|cccccc}
            \toprule
            Method & ImageNet-V2 & ImageNet-A & ImageNet-R & ImageNet-Sketch & ImageNet-ReaL & Avg. \\
            \midrule
            ARM-B  & 72.3        & 31.9       & 47.6       & 35.7            & 87.9           & 55.1 \\
            STAR-B & \textbf{72.5} & \textbf{35.7} & \textbf{49.1} & \textbf{36.4} & \textbf{88.3} & \textbf{56.8} \\
            \bottomrule
        \end{tabular}}
\end{table}

\section{Conclusion}
In this paper, we propose STAR, a long-sequence autoregressive pretraining framework for Vision Mamba. STAR inserts a shared separator before each image in a multi-image sequence and introduces a terminal separator target for the final content cluster. Our study characterizes the separator by its type, value, position, and number. STAR-B achieves 83.5\% EMA top-1 accuracy on ImageNet-1K after 1,600 epochs of pretraining. Its benefits also extend beyond the original classification benchmark: STAR-B raises COCO box and mask AP to 46.84 and 41.45 and improves ARM-B on all five evaluated ImageNet robustness benchmarks, with the mean accuracy increasing from 55.1\% to 56.8\%. Under the evaluated four-image setting and benchmark suite, these results indicate that separator-based long-sequence pretraining improves in-distribution recognition, robustness, and dense-task transfer relative to ARM.

\section{\revise{Limitations and Future Work}}
Our study has several limitations. First, although we extend the evaluation to ImageNet robustness benchmarks and COCO object detection and instance segmentation, the experiments remain centered on ImageNet pretraining and a single MambaMLP backbone scale; broader pretraining corpora, architectures, and application domains remain to be studied. Second, increasing the number of images does not always improve performance, suggesting a practical sequence-length bottleneck in the current backbone. Third, longer packed sequences increase pretraining cost: under the same eight-NVIDIA-A800 setup, STAR requires approximately $1.47\times$ the ARM training time at matched epoch counts. Fourth, although the empirical results demonstrate the effectiveness of cluster-level separators, a complete theoretical explanation of how separator values and positions affect representation learning is still lacking. Finally, dense-task transfer is currently limited to COCO, and semantic segmentation has not yet been evaluated. Future work will investigate more scalable sequence modeling strategies, develop a deeper theoretical understanding of separator-based pretraining, and evaluate STAR across broader datasets, architectures, and dense prediction tasks.

\section*{Declaration of generative AI and AI-assisted technologies in the manuscript preparation process}
During the preparation of this work, the authors used ChatGPT (OpenAI) to improve language and readability and to assist with the organization and presentation of the manuscript. After using this tool/service, the authors reviewed and verified the content, edited it as needed, and take full responsibility for the content of the published article.

\bibliographystyle{elsarticle-num}
\bibliography{main}

\end{document}